# Predictive Scale-Bridging Simulations through Active Learning

**Satish Karra**[1,2,γ], **Mohamed Mehana**[1,γ,*], **Nicholas Lubbers**[3], **Yu Chen**[4], **Abdourahmane Diaw**[5], **Javier E. Santos**[1], **Aleksandra Pachalieva**[1], **Robert S. Pavel**[6], **Jeffrey R. Haack**[7], **Michael McKerns**[7], **Christoph Junghans**[6], **Qinjun Kang**[1], **Daniel Livescu**[7], **Timothy C. Germann**[8], and **Hari S. Viswanathan**[1]

[1]Computational Earth Science Group, Earth and Environmental Sciences Division, Los Alamos National Laboratory, NM, USA 87545
[2]Environmental Molecular Sciences Laboratory, Pacific Northwest National Laboratory, Richland, WA, USA 99354
[3]Information Sciences Group, Computer, Computational and Statistical Sciences Division, Los Alamos National Laboratory, Los Alamos, NM, USA 87545
[4]Department of Mechanics and Aerospace Engineering, Southern University of Science and Technology, Shenzen, China 518055
[5]RadiaSoft LLC, 6525 Gunpark Dr., Suite 370-411, Boulder, CO, USA 80301
[6]Applied Computer Science Group, Computer, Computational and Statistical Sciences Division, Los Alamos National Laboratory, Los Alamos, NM, USA 87545
[7]Computational Physics and Methods, Los Alamos National Laboratory, Los Alamos, NM, USA 87545
[8]Physics and Chemistry of Materials Group, Theoretical Division, Los Alamos National Laboratory, Los Alamos, NM, USA 87545
*Corresponding author: Mohamed Mehana, mzm@lanl.gov [γ]These authors contributed equally

## ABSTRACT

Throughout computational science, there is a growing need to utilize the continual improvements in raw computational horsepower to achieve greater physical fidelity through scale-bridging over brute-force increases in the number of mesh elements. For instance, quantitative predictions of transport in nanoporous media, critical to hydrocarbon extraction from tight shale formations, are impossible without accounting for molecular-level interactions. Similarly, inertial confinement fusion simulations rely on numerical diffusion to simulate molecular effects such as non-local transport and mixing without truly accounting for molecular interactions. With these two disparate applications in mind, we develop a novel capability which uses an active learning approach to optimize the use of local fine-scale simulations for informing coarse-scale hydrodynamics. Our approach addresses three challenges: forecasting continuum coarse-scale trajectory to speculatively execute new fine-scale molecular dynamics calculations, dynamically updating coarse-scale from fine-scale calculations, and quantifying uncertainty in neural network models.

## Main

Recent breakthroughs in machine learning (ML), including the stunning successes of AlphaZero and AlphaGo, have demonstrated tremendous potential for transforming scientific computing.[1–3] By way of analogy, we identify three crucial pieces needed for ML-enabled scale-bridging: (1) advances in algorithms such as cooperative deep neural networks (NNs) that emulate fine-scale simulations within prescribed error bounds, while incorporating known physical constraints and domain expertise, (2) active learning (AL) methods to generate large and targeted datasets intelligently, and (3) massive computational resources, including ML-optimized processors. (For instance, the fastest supercomputers in the United States in 2011, Summit and Sierra at Oak Ridge and Lawrence Livermore National Laboratories, respectively, include NVIDIA Voltas with both tensor and GPU cores, although few scientific computing workloads still utilize them 10 years later). Integrating these advances in ML algorithms and hardware within a scale-bridging framework would transform scientific computing applications by learning to represent the subscale physics which is often the weakest link in the current modeling capabilities.[4,5]

Designing effective methods for multi-scale simulation is a longstanding challenge.[6,7] Without a clear separation of scales, the difficulty becomes tremendous. Using ML to emulate fine-scale models and learn an effective fine/coarse scale-bridging protocol is a very exciting prospect. This allows us to advance state of the art for ML beyond sequential training and inference and facilitate scale-bridging through novel techniques. AL is a special case of semi-supervised ML in which a learning algorithm can interactively use the fine-scale model to obtain the desired outputs at new data points, making it ideal for concurrent scale-bridging. Optimally, an AL procedure should dynamically assess uncertainties of the ML model, query new fine-scale simulations as necessary, and use the new data to incrementally improve the ML models. In order to achieve this goal, three grand ML challenges need to be addressed: (1) How do we best emulate the coarse-scale dynamics to anticipate and speculatively execute relevant fine-scale simulations? NNs and heuristics can be used to estimate the trajectory of coarse-scale states. (2) How do we best dynamically update NNs as new training data is generated? Careful sub-sampling of existing data in addition to newly generated data can be used to retrain the NNs and update the uncertainties. Tensor cores are well suited for this step; their lower precision computations are an acceptable trade-off for the massive gains in performance, since the NN predictions are inherently noisy. (3) How do we estimate ML uncertainties? One possible solution is a cooperative learning approach using a pair of NNs, one for prediction and the second for uncertainty computation. However, one traditional drawback of NNs is that they are unstable outside their calibration range. Thus, alternatively, it may be posited that quantified uncertainties are not required, since their ultimate purpose is to decide which new fine scale simulation results are most valuable. By automatically determining whether a NN is used outside its calibration range, new data can be generated that extends its range when necessary.

In this paper, we select two exemplar applications that exercise different parts of this AL-enabled scale-bridging approach. Inertial confinement fusion (ICF) interfaces are very dynamic and here we demonstrate how we address the ML aspects (1) and (2) for the ICF application. Since nanotransport in shale is less dependent on coarse-scale problem but is plagued with uncertainty since the subsurface is opaque, we demonstrate incorporating uncertainty in NN calculations and addressing ML challenge (3). Our overall research goal is to transform NNs for scientific computing by unlocking their enormous potential to learn complex hierarchical physics from big data with training times that scale linearly with dataset size.

# Results

We present an integrated, multi-disciplinary framework (Fig. 1) where coarse-scale model, Lattice Boltzmann Method (LBM) or kinetic multi-ion Vlasov-Bhatnagar-Gross-Krook (Multi-BGK) model in our two applications, respectively, is divided into subdomains that are partitioned across nodes. When the coarse-scale model is insufficient to advance the simulation, the Active Learner NN either generates an approximation from existing results or performs a new Fine-Scale Molecular Dynamics (MD) computation, depending on the estimate of the Uncertainty Quantifier (UQ) NN. The new result is added first to the local database and then to an eventually consistent distributed database, in a trade-off between duplicated computing and communication. The computational realization of these algorithms requires several aspects of next-generation architecture/algorithm design, including the reliance on an asynchronous task-based computing paradigm (and enabling programming and execution model technologies) and the use of tensor cores for the NN training and inference. All of these aspects are required to fully exploit the massive concurrency and heterogeneity of future exascale computing platforms. While our AL algorithms and computational framework are generally applicable, they are developed and demonstrated for two specific applications where molecular-scale physics impacts the results of continuum-scale models, as illustrated in Fig. 2 and described below in more detail.

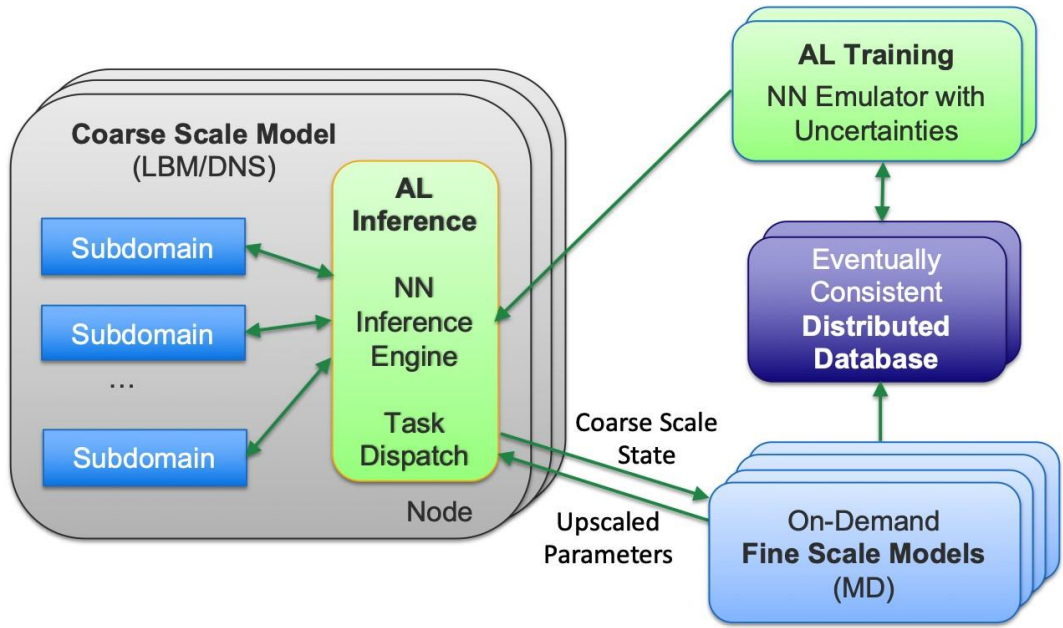

**Figure 1.** Scale-bridging framework, with coarse- and fine-scale models, database, and active learning tasks. The latter utilize GPUs (or other accelerators), both for the infrequent but expensive NN training (of upscaled parameters and their uncertainties) and the frequent NN inference.

ICF targets consist of a small quantity of DT fuel surrounded by a heavier shell that compresses the fuel enough to enable thermonuclear fusion.[8] However, in practice, there is a significant degradation in the compression due to hydrodynamic instabilities and interfacial mixing, as well as heat loss from the central hot spot. The overlap between the flow and molecular scales results in purely hydrodynamic simulations missing important effects such as non-local transport. Nevertheless, due to resolution constraints, state of the art for ICF problems uses meshes coarse enough to fit on current supercomputers and uses non-physical numerical diffusion and relatively ad hoc models to approximate molecular effects.[9] Recent attempts to overcome such crude approximations either address the flow scales with more accurate continuum solvers[10], higher resolution[11], or account for high temperature molecular effects using electron quantum descriptions.[12] Such simulations have highlighted the importance of flow instabilities and plasma transport for the hot spot dynamics. However, more accurate early-time simulations are needed to identify the exact structure of the flow perturbations. For example, our preliminary results of the Marble experiments described below show a complex flow evolution during pre-heat, with self-generated shocks temporarily halting the mixing. Similarly, for late-time mixing, orbital free density functional theory (OF-DFT) calculations[12] show that common transport models can have significant errors and may need to be locally updated.[9] Non-local transport may also require different continuum-level descriptions and not be amenable to off-line property evaluations. The accuracy of MD has been demonstrated as an efficient computational tool to probe the microscopic properties of high-energy density experiments.[13] MD was used to validate various microscopic models that were implemented in an ICF hydrodynamics code: equilibration phenomena, structure, and radiative processes.[14] However, MD cannot currently be performed over time and length scales relevant for ICF experiments as it becomes very expensive at higher temperatures. For instance, a state-of-the-art massively parallel MD simulation of a 2 micron capsule-shell ICF interface took a few weeks to simulate 10 ps, which is only about 0.2-0.5% of the total implosion time, underscoring the need for a new multi-scale approach.

In the second application, the majority of the pore space in rock formations (e.g., shale) relevant to subsurface energy and water resources exists in pores of 100 nm or less, where continuum models often break down.[15] Quantitative predictions of transport in nanoporous media are critical to optimize hydrocarbon extraction from low permeability shale formations through hydraulic fracturing. We have shown that methane production is vastly underestimated because Knudsen diffusion, one of several molecular (subnanoscale) effects, is not accounted for in coarse-scale (nanoscale) Navier-Stokes (NS) models. Knudsen diffusion can result in a 10-100x increase in the bulk property permeability.[16] Intuitively, there are many more interactions between fluids and confining pore walls at these small pore sizes (10-100 nm) that could result in emergent phenomena, affecting methane extraction.[17] While the effects of microscopic random fluctuations on coarse-grained models using hybrid algorithms have been studied for a few problems including linear diffusion[18] and the inviscid Burgers[19], Navier-Stokes[20], and Ginzburg-Landau[21] equations, direct atomistic-continuum scale-bridging models have been challenging to implement. This is primarily due to the disparate length and timescales resulting in inaccurate information transfer across scales and poor computational performance.[22] Although there are some studies on hybrid atomistic-continuum models (including coupling LBM and MD) for dense fluids, these focus on very simple systems, e.g., single-phase laminar flow past or through a carbon nanotube.[23] Scale-bridging simulations for nanoporous flow relevant to subsurface applications are essentially non-existent or limited to studies on off-line use of molecular simulations to generate models for LBM (i.e., "sequential scale-bridging")[17], ignoring any overlap in scales.

**Inertial Confinement Fusion**

Inertial confinement fusion experiments are fundamentally multi-scale in nature. When modeling these experiments, quantum molecular dynamics and similar models collect data on the dynamics and structure of processes occurring at microscopic scales.[24] However, ICF experiments and their results evolve on macroscopic scales in space and time that are much larger than can be simulated by microscopic models; an accurate understanding of the connection between the expensive microscale physics and the experimental observables is needed. Large scale properties are crucially affected by microscale information such as equations of state and ionic and electronic transport coefficients. This additional closure information is provided by the microscale model, which is communicated to the macroscale via, e.g., tables, formulas, etc. For more accurate closure information, a large number of multi-scale techniques have been developed to bridge these scales[24], and while they perform well for relatively simple systems, the sheer number of fine-scale calculations needed for more realistic problems can quickly become impractical without some additional guidance.[25]

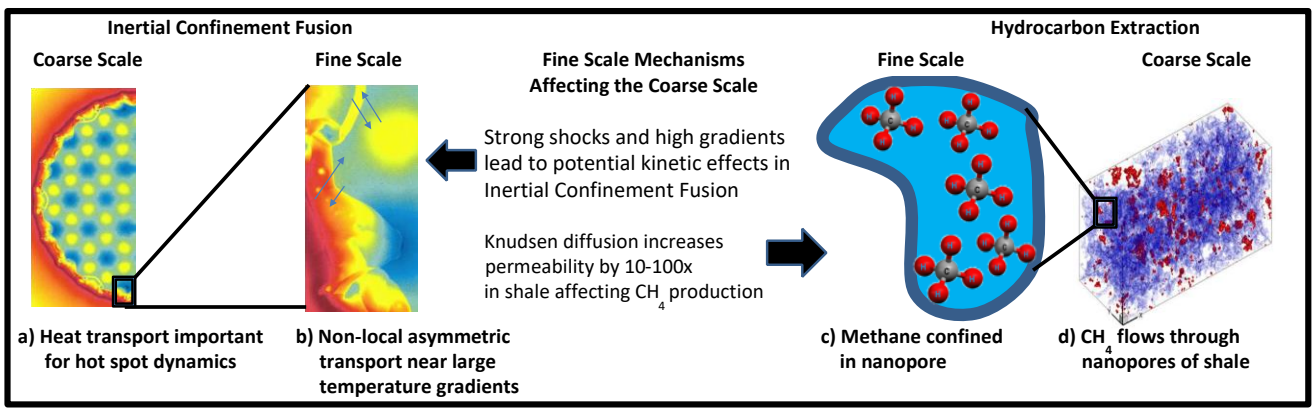

**Figure 2.** Our machine learning based scale-bridging framework for inertial confinement fusion and shale nanoconfinement applications.

One of the most challenging features to measure in ICF experiments is the atomic mixing of materials in dense plasmas. In particular, this issue is relevant to MARBLE-like ICF experiments, in which ICF targets are filled with engineered deuterated foam.[26] The foam's pores are made to a specified size and filled with a gas containing tritium; the resulting DT fusion yield of the initially separated reactants in the experiment gives a measure of the amount of mixing that has occurred. To model the atomic mixing, we use a multi-ion Vlasov-Bhatnagar-Gross-Krook kinetic model, for which the key closure is the collision frequencies between species.[27] Standard models for these frequencies rely on weak coupling assumptions, which significantly simplify the analysis but may not be valid in the pre-heat regime of an ICF capsule implosion. Instead, we use MD to compute multi-species diffusion rates via Green-Kubo relations[28], which are then used to define the collision rates in the kinetic model. The GLUE code interfaces the kinetic simulation, the MD simulation, the surrogate model, and the computing platform to perform the overall multi-scale simulation.

We demonstrated this coupling using the kinetic code Multi-BGK and MD code LAMMPS to study mixing at an ArgonDeuterium interface.[29] To reduce the cost of gathering information from MD, we used active learning to train neural networks on MD data obtained by randomly sampling a small subset of the parameter space. We constructed our dataset as a random set of a Latin hypercube experiment design containing 1100 points in the five-dimensional input parameter space. In order to capture the expected conditions from MARBLE, we focus on material densities in the range $10^{22}$ cm$^{-3}$ ≤ $n$ ≤ $10^{25}$ cm$^{-3}$, and a plasma temperature $T$ in the range 50eV ≤ $T$ ≤ 150eV. Our model performance on test MD data is shown in Fig 3 (top row). Points are shown in colored using the quality flag $s_i$, which allows selecting points where the model confidence is low and requesting new MD calculations.

We applied this method to a plasma interfacial mixing problem relevant to a warm dense matter, showing considerable computational gains compared to the full kinetic-MD approach. This approach enables probing Coulomb coupling physics across a broad range of temperatures and densities inaccessible with current theoretical models. The surrogate model gives a result within an acceptable measure of accuracy compared to an all-MD simulation, at much less expense. When the material is artificially heated outside the range of the data used to initialize the surrogate model, the learner framework can successfully update itself; see Figure 3(bottom row).

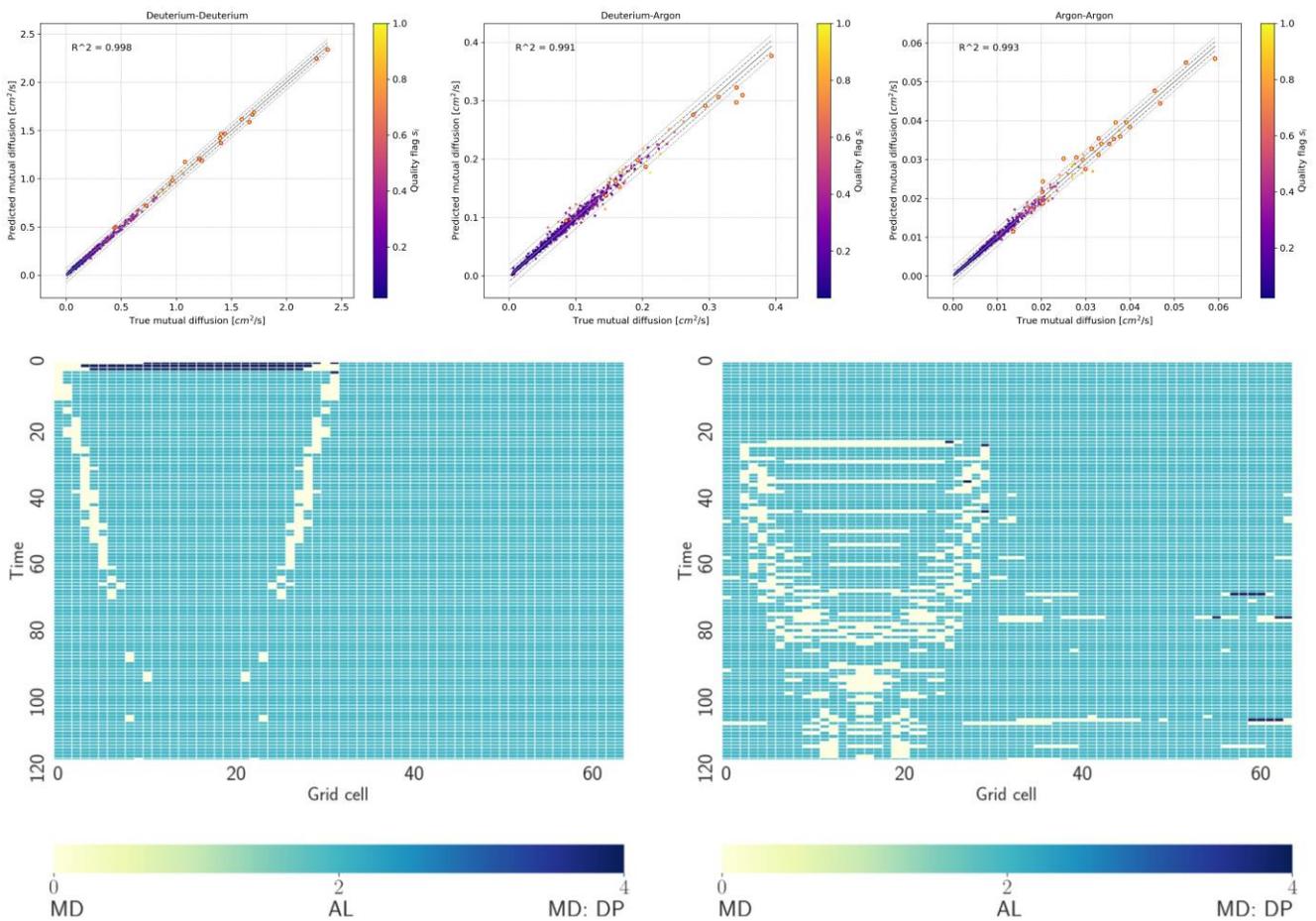

**Figure 3.** Top row: We show the performance of the ML surrogate model for all the mutual diffusion points in the test dataset, which were generated by randomly sampling 1100 points from a 5-D input parameter space. From left to right: Deuterium-deuterium, Deuterium-Argon, and Argon-Argon mutual diffusions. The points selected for the MD verification are circled in red. Bottom row: Locations of MD calls during an interfacial mixing simulation. White = MD, Blue = surrogate model, Black=database duplicate. The simulation on the left used a uniform temperature profile, while the case on the right is artificially heated to push the state of the system away from the initial training data used for the surrogate model.[29]

We proposed an alternative method to evaluate the diffusion coefficients required to close the kinetic model, which applies modern data science tools to predict the thermodynamic factor over multi-species phase space.[30] The thermodynamic factor measures ideal or non-ideal mixing, which can be used with species self-diffusion coefficients to generate the inter-species diffusion terms needed by the kinetic model. These inter-species terms are relatively more expensive to compute than the self-diffusion terms, especially in cases where there are trace amounts of one species in a mixture. This improvement allows for a significant decrease in the amount of time needed by an MD call to return relevant data to the kinetic model and speeds up the multi-scale simulation. Because fine-scale simulation is so expensive relative to coarse-scale simulation, it is essential to choose data points as prudently as possible to maximize the improvement of the surrogate model, which also improves the overall multi-scale performance as it reduces the number of requests that the surrogate cannot answer.

To this end, a sampling workflow is provided that scores surrogates against newly evaluated data and has an additional optimization loop that attempts to find a surrogate where its validity has converged to an optimum. In Diaw et al.[31], we assessed the performance of different sampling strategies in learning asymptotically valid surrogates for benchmark functions. We compared a more traditional sparse sampling that finds and samples at the least-populated points each draw to an optimizerdirected approach where each initial draw is used as starting point for an optimizer that runs to the termination. We used a metric based on the error in the surrogate's predicted value versus truth, averaged over the entire response surface. Our methodology produced surrogates that are, on average, of high quality across the whole parameter range, regardless of the sampling strategy used. We found that, given that one does some upfront work to tune the optimizer's termination conditions for the given problem, the optimizer-directed approach can be more efficient than a traditional sampling approach. Importantly, we also noted that the metric we used did not guarantee the quality of the surrogate *in the neighborhood of the turning points*. We found that an optimizer-directed approach is much better at minimizing the model error in the neighborhood of the turning points (see Figure 4), even when the metric does not call for that explicitly. Conversely, a traditional sampling strategy, which is blind to the response surface, generally demonstrated a much larger misfit near the turning points. Thus, using a metric that judges the quality of the surrogate by the misfit at the turning points should produce high-quality surrogates with an optimizer-directed approach with even greater efficiency.

The critical points of the response surface are the key points to find to guarantee the long-term validity of the surrogate as new data is collected, and are generally where the interesting physics occurs. To demonstrate this, we applied our methodology to two test problems. We showed that we could efficiently learn surrogates for the equation of state calculations of dense nuclear matter, yielding excellent agreement between the surrogate and model both across the parameter range and specifically in the region, which includes physically-relevant features, such as a phase transition. We also showed that our methodology can produce highly-accurate surrogates for radial distribution functions, expensive molecular dynamics simulations, neutral and charged systems of several dimensions across an extensive range of thermodynamic conditions.

**Nanoconfinement in Shale**

Shale reservoirs have redefined the energy landscape in the world.[32] Currently, shale reservoirs account for 70% of the total US gas production and 60% of the entire US oil production.[33] However, their characteristics significantly differ from conventional reservoirs. Among these characteristics is the nanoconfinement, where the fluid/pore wall interactions dominate the fluid behavior.[34] In shale, approximately 80% of the pores is less than 50 nm. At this scale, surface phenomena such as adsorption and slip control the equilibrium and transport properties of the fluid.[35] However, most of the current simulators do not account for these effects.[16]

Given the challenges of the nanoscale experiments, we augmented the limited experimental data with molecular simulation to reveal the confinement effects. MD simulation has become a necessary research technique to conduct computational experiments, giving us an ever-increasing ability to predict matter's physical and chemical properties and probe experimentally inaccessible areas such as confinement. However, the computational cost of molecular simulation is prohibitive to studying natural systems. Therefore, scale-bridging techniques become necessary to upscale the molecular simulations insights to continuum pore-scale simulations.[5]

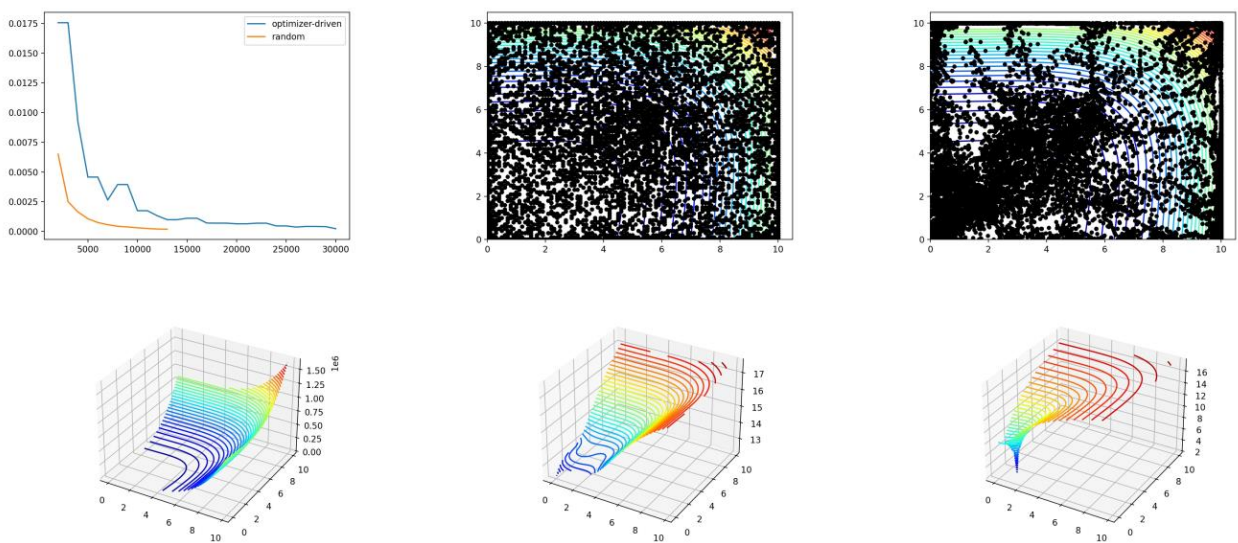

**Figure 4.** Candidate surrogates for the 8-dimensional Rosenbrock function, learned with a thin-plate RBF estimator, using a SparsitySampler from *mystic* and a validity based on average model error. Top row, left: test score per iteration with optimizer-directed sampling in blue, and pure systematic random sampling in orange. Top row, center: model evaluations sampled with the random sampling strategy. Top row, right: optimizer-directed sampling, using an ensemble of Nelder Mead solvers in the default configuration. Bottom row, left: surrogates produced with either sampling approach appear identical to the truth. Bottom row, center: log-scaled view of surrogate from random sampling near the global minimum. Bottom row, right: log-scaled view of surrogate from optimizer-directed sampling near the global minimum, identical to the truth. While pure systematic random sampling converges faster than when the optimizer's default configuration is used, optimizer-driven sampling generally provides a more accurate surrogate near the critical points.

We start with MD simulation to derive the physics of confined fluids such as adsorption, slip length, and diffusivity.[15] After that, we adopt deep-learning techniques to develop scale-bridging workflows to calibrate pore-scale simulations to account for the confinement effects. Our crucial innovation relies on designing and training neural network models to enable active learning and uncertainty quantification. Figure 5 presents one of these innovations, modeling the adsorption behavior in nanopores. We overcome the computational burden of fine- and coarse-scale simulations using efficient emulators. These emulators were used afterward to train an upscaler that maps the fine-scale simulations with the coarse- scale simulations. Although the upscaler could be trained directly from the fine- and coarse- scale simulations, our trained emulators, enabled speculatively to execute new fine-scale calculations based on coarse-scale trajectory forecast. It is worth noting that molecular dynamics consider the molecular and discrete nature of the matter, unlike our continuum-scale simulations; consequently, we adapted an indirect coupling scheme to match the adsorption phenomena in this case.

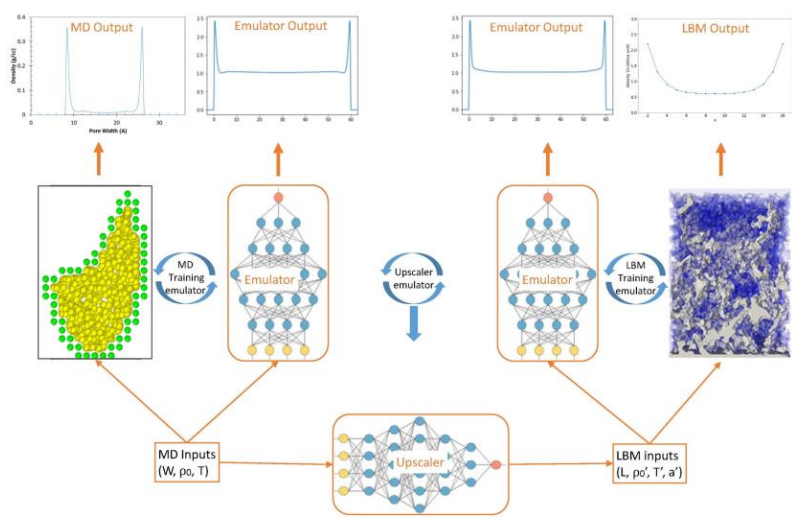

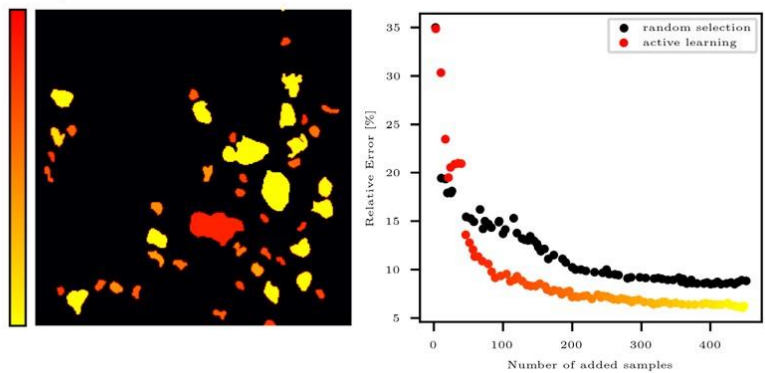

**Figure 5.** Top is a detailed workflow to include nanoconfinement effects in pore-scale simulations. We train Deep Neural Network emulators using molecular dynamics (MD) and Lattice Boltzmann method (LBM) simulations of pore profiles under a range of conditions. These emulators are later used to train an upscaler that updates the LBM parameters to include the confinement effects (for instance, adsorption in this case). Bottom right is the performance of uncertainty-guided active learning strategy. This approach reaches error targets faster than conventional approaches. Bottom left shows high and low uncertainty pores from fine-scale emulator predictions. The active learning strategy uses uncertainty in the neural networks to guide the training process.

We applied our workflow to a simple silt pores where MD simulations were used to quantify the adsorption characteristics and inform the pore-scale simulations. A promising approach is to leverage deep learning techniques to couple the MD simulations with the pore-scale simulations.[36] Moving to more complex pore structures, we adopted an active learning framework to accelerate neural network training.[37] We quantified the uncertainty in the neural networks by identifying the areas of mismatch among randomly-initialized and similarly-trained neural networks. This allows to dynamically update and augment the fine-scale database with highly-uncertain data points. We successfully reached the accuracy of conventional approaches using only 10% of the dataset through uncertainty-based training. This work demonstrates a workflow to build the fine-scale emulators using an active learning approach and utilizing an uncertainty-guided data collection strategy. We extended the work scope to model the transport properties using reinforcement learning.[38] We used a new convolutional autoencoder model for emulating direct numerical simulation of fluid flow through porous materials, where we included novel methods to impart knowledge of the Navier-Stokes and continuity equations. In addition, we increase the learning stability, reduce over-fitting, and promote human interpretability of the neural network layers by using hierarchical regularization of the loss function. Our new model is used to emulate coarse- and fine-scale simulations of fluid flow through pore-structures, where orders of magnitude speedup of computation times is realized.

## Discussion

The critical aspect of bridging scales is usually some form of constitutive equation that specifies the relationship between macro-scale observables and macro-scale responses as a result of underlying micro-scale interactions that are not otherwise specified by the macro-scale equations. This is often termed closure. Empirical constitutive relations are also often termed surrogate models. In near-equilibrium scenarios that are well described by the macroscopic theory, very simple constitutive relationships are satisfactory, for example, use of a constant diffusion coefficient or fluid viscosity. However, specific models need to be developed to capture the physics in extreme conditions such as ICF and nanoporous media. For example, BGK[39] models suitable for ICF require closure information to specify the typical collision rates of particles of each species with each other. Simple models that capture scaling behavior in a perturbative context (e.g., weakly coupled plasma) are not completely sufficient to cover contexts where the underlying physics changes, such as in the strong-coupling regime.

To capture the behavior of microscopic physics in a complex system, researchers have often built by hand surrogate models that capture the behavior from data describing the underlying microscopic processes using empirical functional forms with adjustable parameters. This process does not naturally treat many variables simultaneously, and can be laborious for a researcher to find an appropriate functional form. Additionally, one may not know ahead of time, what phase space might be encountered in the macroscopic system, resulting in an iterative need to expand surrogate models to cover more and more data.

The rise of machine learning for automation is a natural fit for constructing surrogate models in complex conditions.[40,41] Historically, machine learning has developed out of pattern-recognition-based endeavors to make progress in artificial intelligence. More or less, by definition, this means that machine learning workflows are geared toward automation. Models have been developed for a variety of tasks, many of which are possible or even simple for a human to accomplish but difficult to directly program solutions for, as a result of the ample space of possible inputs and the exponentially large space of possible functions on these inputs. As a result, machine learning methods are capable of searching this high-dimensional space of functions for useful empirical models that can be evaluated quickly by computers. Once a model is trained, calls to extract information can be very cheap and can often be executed millions of times at a negligible computational cost. They have been scaled up to treat massive datasets.[41] This results in models that do not assume the apriori functional form of the upscaling functions from fine-scale (MD) inputs to coarse-scale (LBM or kinetic theory) parameters. Scale-bridging results in numerous small simulations which can be treated as a dataset for machine learning. Like the many fine-scale simulations in scale-bridging, many machine learning methods can be parallelized to reduce training time.

We developed an AL-based framework that efficiently and accurately captures the fine-scale physics and can be used to bridge fine and coarse scales while reducing the overall computation time by orders of magnitude, depending on the problem and required accuracy. We also provided broadly useful advances in scientific ML: quantifying the uncertainty in NN models, forecasting the coarse-scale trajectory to speculatively execute new fine-scale calculations, and dynamically updating an NN emulation model as new data is obtained.

We demonstrated our approach for hydrocarbon extraction, where we tightly coupled MD and LBM simulations of methane adsorption and transport in nanopores. In addition, we extended our approach to the most critical nanoscale physicochemical processes by training the AL on MD calculations of parameters required by LBM, accounting for the effects of nanoscale confinement on phase behavior. On the other hand, we developed a consistent framework for ICF to bridge continuum and MD scale simulations that remains computationally feasible while also dramatically increasing the accuracy of the calculations. Our work demonstrated the potential of active learning techniques to develop efficient and accurate scale-bridging workflows for various applications and fields.

## Methods

We present the details of our approach and workflow.

*Neural Networks and scale-bridging*

Machine learning approach to scale-bridging has been traditionally addressed using Gaussian Process regression (GPR). GPR constructs a regression model based on a kernel function.[42] The kernel function specifies the correlation between the response function of data points; if the kernel is large, the two points are assumed to have nearly the same response. In this sense, the kernel can be thought of as a type of distance or similarity function. GPR emulators have proven remarkably successful in a number of contexts such as optimal design, simulating radiative transfer and modeling fluid flow.[43] However, a crucial limitation of GPR is the cost of building a model is associated with a matrix inverse problem, which scales like $O(N^3)$ for $N$ data points, and thus does not scale effectively as a dataset grows in size. Another drawback of GPR and other kernelized models is finding a suitable kernel as the number of features grows larger; this crucial ansatz often suffers under the curse of dimensionality, and so it can be hard to build accurate GPRs in high-dimensional spaces. In addition, previous efforts to apply kriging-based adaptive sampling techniques to multi-scale materials applications found that neither a global GPR model, nor patching together several local GPR models, worked well.[44]

To alleviate the limitations of the GPR approach, we utilized machine learning and, in particular, Deep Neural Networks (DNN) instead. The main advantages of using neural networks in comparison to traditional scale-bridging methods are summarized below:

- The training can be performed over a multivariable parameter space simultaneously. This allows for handling large datasets and simultaneously incorporates many simulation properties.

- Our NN approach does not depend on the dimensionality of the system. Currently, our framework solves 1D pores. However, it can be generalized for two- and three-dimensional pore systems.

- Unlike the traditional scale-bridging models, the upscaled functions do not need to assume the functional form of the parameters a priori. The only assumption is that the functions can be generated by a neural network.

- Our framework can be extended to enforce physical constraints such as flow conditions, equation of state modifications, phase transition characteristics, and conservation low such as mass and momentum (current).

- The DNN models also take advantage of fast automatic differentiation algorithms for computing gradients, which are available within DNN libraries.[45] Using automatic differentiation, the emulator and the upscaler networks are trained faster and with less laborious deriving and coding numerical finite-difference gradients.

In indirect scale-bridging, the fine-scale inputs do not need to be explicitly matched to the fine-scale inputs. Instead, an additional upscaler is trained based on emulators defined on the fine-scale and the coarse-scale input data. Since both emulators are defined on the basis of the respective input space of the collected data, the upscaler solves the problem implicitly. The network incorporates the properties of many simulations simultaneously because it operates on multivariate parameter space instead of optimizing upscale parameters on a point-by-point basis. We use this strategy in the nanoporous media transport application, where an absorption parameter obtained from temperatures, pore width and density profiles.

*Forward UQ and active learning*

It is essential to ensure that the surrogate model captures the behavior of MD simulations in the entire space of potential MD simulations. Collecting MD data, as mentioned earlier, is very expensive, and often there is no apriori knowledge of the needed MD calculations. We reduce this cost by applying active learning, which includes the following steps: (1) perform some MD calculations, in parallel, ahead of time to build a surrogate model. The input parameters of the MD runs are chosen randomly by sampling the parameter space; (2) while querying the surrogate model, we request an UQ model, which predicts the uncertainty of the surrogate model; (3) if the UQ model reports that the prediction carries a large uncertainty, new MD simulations are spawned to collect ground-truth data.

The surrogate model is then periodically retrained so that it accounts for all available MD data. This ensures that the surrogate model is flexible and can adapt to the new macro-scale conditions on the fly. In this case, using ML and UQ is expected to increase the accuracy of the surrogate model in comparison to approximations and it reduces the computational time spent in MD simulations by collecting only the necessary data automatically. We build an ensemble for query-by-committee uncertainty training each network to a random subsection of the dataset. A random 10% of the data is withheld for calibration. Each ensemble produces a different model since the initialization of the network weights and the data split is random. We calculate the network performance of each model using its calibration data. The model delivers good results if the coefficient of determination is $R^2 >= 0.7$, otherwise, it is rejected. This procedure is executed until $n_{ensemble} = 5$ networks have been successfully trained.

*Active learning and asymptotic validity*

Recently, we used active learning to generate surrogates of fine-scale materials response[29,36], while Roehm et al.[46–48] used kriging to construct surrogates of stress fields and communicate the results to a fine-scale code that solves the macro-scale conservation laws for elastodynamics. Noack et al.[49] used a similar kriging-based approach to construct surrogates for autonomous x-ray scattering experiments. None of the above approaches attempt to ensure surrogates are valid on future data and thus provide no guarantee on the validity of the surrogate. Instead, whenever a learned surrogate is evaluated, an uncertainty metric is also evaluated to determine if and where new fine-scale simulations should be launched. For example, Noack et al.[49] uses a genetic algorithm to find the maximum variances for each measured data point, then draws new samples from a distribution localized around the solved maximum. Passive approaches to validity such as these assume the expensive model is always available and may have performance significantly hampered by the need to frequently request new expensive model evaluations throughout the life of the surrogates.

We present a new online learning methodology to efficiently learn surrogates that are *asymptotically* valid with respect to any future data. We use asymptotically valid to indicate that while we do not have a formal proof of validity versus future data, there is strong evidence to support our claim of at least approximate validity with respect to future data under some light conditions. More specifically, we conjecture that the minimal data set necessary to produce a highly accurate surrogate is composed of model evaluations at all critical points on the model's response surface. Our conjecture comes with the condition that the selected surrogate class has enough flexibility to reproduce the model accurately. Hence, it is beneficial to use a radial basis function (RBF), multilayer perceptron (MLP), or other similar estimators with universal function approximation capability when training a surrogate for the model's response surface.

Our methodology has three key components: a sampling strategy to generate new training and test data, a learning strategy to generate candidate surrogates from the training data, and a validation metric to evaluate candidate surrogates against the test data. For implementation, we leverage *mystic*[50], an open-source optimization, learning, and uncertainty quantification toolkit. *mystic* has over a decade of use in the optimization of complex models, including using uncertainty metrics to optimally improve model accuracy and robustness and to increase the statistical robustness of surrogates. Recent developments have included configurable optimizer-directed sampling, which we use in combination with online learning to train surrogates for accuracy with respect to all future data.

Our general procedure for producing a valid surrogate starts with an initial surrogate built by training against a database of existing model evaluations. The surrogate is then tested against a validity condition (i.e., a quality metric, such as the expected model error is less than a given tolerance). If the surrogate is deemed to be valid after testing, execution stops. Otherwise, the surrogate is updated by retraining against the database of stored model evaluations, where the surrogate is validated with a fine-tuning of surrogate hyperparameters against a quality metric. If iterative retraining improves the surrogate, it is saved. Otherwise, we sample new model evaluations to generate new data. The process repeats until testing produces a valid surrogate.

Our procedure for producing a valid surrogate is extended for asymptotic validity by adding a *validity convergence condition* to be called after the surrogate is deemed *test* valid. Thus, instead of stopping execution when the surrogate is *test* valid, we merely complete an iteration toward validity convergence. If not *converged*, we trigger a new iteration by sampling new data and continue to iterate until the surrogate validity has converged. This iterative procedure is more likely to generate a surrogate that is valid for all future data, as we are testing for convergence of validity against unseen data over several iterations. When the database of model evaluations is sparsely populated, we expect that any new data will likely trigger a surrogate update. Note that adding new data to the database does not ensure that a surrogate will become more valid; however, our conjecture is that as we sample more of the critical points on the model's response surface, our ability to learn a valid surrogate improves. To reduce the number of sampling iterations required, we can increase the number of samplers used in each iteration.

We focused on how the choice of sampling strategy impacts efficiency in producing an asymptotically valid surrogate, with validity defined by the evolution of the surrogate test score. We compared optimizer-directed sampling versus more traditional random sampling in the efficiency of learning the response surface, where, fundamentally, both approaches utilize an identical first draw. In 'optimizer-driven' sampling, the other draw uses first draw members as starting points for the optimizer instead of repeating the same probability sampling used in the first draw. This approach follows a snowball sampling of points for each first draw member, with the corresponding optimizer directing the sampling toward a critical point on the response surface. When the termination condition of the optimizer is met, probability sampling is used as in the first draw to generate a new starting point for a new optimizer and again continues until termination.

While the sampling and learning components of our workflow are fundamentally independent and are able to run asynchronously, they are linked through the database of stored model evaluations. The data points generated by the sampler are populated to the database, while the learning algorithm always uses the data contained in the database when new training is requested. If there were no concerns about minimizing the number of model evaluations, we could have samplers run continuously, feeding model evaluations into the database. However, we include sampling as part of the iterative workflow, as described above, to minimize the number of model evaluations.

**1 Data availability**

The datasets used and/or analysed during the current study available from the corresponding author on reasonable request

**Acknowledgments**

This work was supported by the Laboratory Directed Research and Development program at Los Alamos National Laboratory (LANL) under project number 20190005DR. LANL is operated by Triad National Security, LLC, for the National Nuclear Security Administration of US Department of Energy (Contract No. 89233218CNA000001). We also thank the LANL Institutional Computing Program and CCS-7 Darwin cluster for computational resources. SK thanks Environmental Molecular Sciences Laboratory for support. Environmental Molecular Sciences Laboratory is a DOE Office of Science User Facility sponsored by the Biological and Environmental Research program under Contract No. DE-AC05-76RL01830.

**References**

1. Scherer, M. *et al.* Machine learning for deciphering cell heterogeneity and gene regulation. *Nat. Comput. Sci.* 1, 183–191 (2021).

2. Nguyen, N. D., Huang, J. & Wang, D. A deep manifold-regularized learning model for improving phenotype prediction from multi-modal data. *Nat. Comput. Sci.* 2, 38–46 (2022).

3. Scherbela, M., Reisenhofer, R., Gerard, L., Marquetand, P. & Grohs, P. Solving the electronic schrödinger equation for multiple nuclear geometries with weight-sharing deep neural networks. *Nat. Comput. Sci.* 1–11 (2022).

4. Yucel Akkutlu, I. & Fathi, E. Multiscale gas transport in shales with local kerogen heterogeneities. *SPE journal* 17, 1002–1011 (2012).

5. Mehana, M., Kang, Q., Nasrabadi, H. & Viswanathan, H. Molecular modeling of subsurface phenomena related to petroleum engineering. *Energy & Fuels* 35, 2851–2869 (2021).

6. Tadmor, E. B. & Miller, R. E. *Modeling materials: continuum, atomistic and multiscale techniques* (Cambridge University Press, 2011).

7. Nagarajan, A., Junghans, C. & Matysiak, S. Multiscale simulation of liquid water using a four-to-one mapping for coarse-graining. *J. Chem. Theory Comput.* 9, 5168–5175, DOI: 10.1021/ct400566j (2013). PMID: 26583426, https://doi.org/10.1021/ct400566j.

8. Lindl, J., Landen, O., Edwards, J., Moses, E. & team, N. Review of the national ignition campaign 2009-2012. *Phys. Plasmas* 21, 020501 (2014).

9. Rosen, M. *et al.* The role of a detailed configuration accounting (dca) atomic physics package in explaining the energy balance in ignition-scale hohlraums. *High Energy Density Phys.* 7, 180–190 (2011).

10. Weber, C., Clark, D., Cook, A., Busby, L. & Robey, H. Inhibition of turbulence in inertial-confinement-fusion hot spots by viscous dissipation. *Phys. Rev. E* 89, 053106 (2014).

11. Haines, B. M. *et al.* Detailed high-resolution three-dimensional simulations of omega separated reactants inertial confinement fusion experiments. *Phys. Plasmas* 23, 072709 (2016).

12. Stanton, L. G. & Murillo, M. S. Ionic transport in high-energy-density matter. *Phys. Rev. E* 93, 043203 (2016).

13. Graziani, F. R. *et al.* Large-scale molecular dynamics simulations of dense plasmas: The cimarron project. *High Energy Density Phys.* 8, 105–131, DOI: https://doi.org/10.1016/j.hedp.2011.06.010 (2012).

14. Stanton, L., Glosli, J. & Murillo, M. Multiscale molecular dynamics model for heterogeneous charged systems. *Phys. Rev. X* 8, 021044 (2018).

15. Neil, C. W. *et al.* Reduced methane recovery at high pressure due to methane trapping in shale nanopores. *Commun. Earth & Environ.* 1, 1–10 (2020).

16. Chen, L. *et al.* Nanoscale simulation of shale transport properties using the lattice boltzmann method: permeability and diffusivity. *Sci. reports* 5, 1–8 (2015).

17. Li, Z.-Z., Min, T., Kang, Q., He, Y.-L. & Tao, W.-Q. Investigation of methane adsorption and its effect on gas transport in shale matrix through microscale and mesoscale simulations. *Int. J. Heat Mass Transf.* 98, 675–686 (2016).

18. Alexander, F. J., Garcia, A. L. & Tartakovsky, D. M. Algorithm refinement for stochastic partial differential equations: I. linear diffusion. *J. Comput. Phys.* 182, 47–66 (2002).

19. Bell, J. B., Foo, J. & Garcia, A. L. Algorithm refinement for the stochastic burgers' equation. *J. Comput. Phys.* 223, 451–468 (2007).

20. Williams, S. A., Bell, J. B. & Garcia, A. L. Algorithm refinement for fluctuating hydrodynamics. *Multiscale Model. & Simul.* 6, 1256–1280 (2008).

21. Taverniers, S., Alexander, F. J. & Tartakovsky, D. M. Noise propagation in hybrid models of nonlinear systems: The ginzburg–landau equation. *J. Comput. Phys.* 262, 313–324 (2014).

22. Zimon, M. J., Sawko, R., Emerson, D. R. & Thompson, C. Uncertainty quantification at the molecular–continuum model´ interface. *Fluids* 2, 12 (2017).

23. Mohamed, K. & Mohamad, A. A review of the development of hybrid atomistic–continuum methods for dense fluids. *Microfluid. Nanofluidics* 8, 283–302 (2010).

24. Root, S., Cochrane, K. R., Carpenter, J. H. & Mattsson, T. R. Carbon dioxide shock and reshock equation of state data to 8 mbar: Experiments and simulations. *Phys. Rev. B* 87, 224102, DOI: 10.1103/PhysRevB.87.224102 (2013).

25. Dornheim, T. *et al.* The static local field correction of the warm dense electron gas: An ab initio path integral monte carlo study and machine learning representation. *The J. Chem. Phys.* 151, 194104, DOI: 10.1063/1.5123013 (2019).

26. Murphy, T. J. *et al.* Progress in the development of the MARBLE platform for studying thermonuclear burn in the presence of heterogeneous mix on OMEGA and the national ignition facility. *J. Physics: Conf. Ser.* 717, 012072, DOI: 10.1088/1742-6596/717/1/012072 (2016).

27. Haack, J. R., Hauck, C. D. & Murillo, M. S. A conservative, entropic multispecies bgk model. *J. Statisical Phys.* 168, 826–856 (2017).

28. Kubo, R. Statistical-mechanical theory of irreversible processes. i. general theory and simple applications to magnetic and conduction problems. *J. Phys. Soc. Jpn.* 12, 570–586, DOI: 10.1143/JPSJ.12.570 (1957).

29. Diaw, A. *et al.* Multiscale simulation of plasma flows using active learning. *Phys. Rev. E* 102, 023310, DOI: 10.1103/PhysRevE.102.023310 (2020).

30. Rosenberger, D., Lubbers, N. & Germann, T. C. Evaluating diffusion and the thermodynamic factor for binary ionic mixtures. *Phys. Plasmas* 27, 102705, DOI: 10.1063/5.0017788 (2020).

31. Diaw, A., McKerns, M., Sagert, I., Stanton, L. & Murillo, M. Efficient learning of accurate surrogates for simulations of complex systems. *under review Nat. Mach. Intell.* (2022).

32. Hughes, J. D. Energy: A reality check on the shale revolution. *Nature* 494, 307 (2013).

33. Total primary energy supply (tpes) by source, world 1990-2017. https://www.iea.org/data-and-statistics. Accessed: 2020-01-14.

34. Liu, X. & Zhang, D. A review of phase behavior simulation of hydrocarbons in confined space: Implications for shale oil and shale gas. *J. Nat. Gas Sci. Eng.* 68, 102901 (2019).

35. Mehana, M., Kang, Q. & Viswanathan, H. Molecular-scale considerations of enhanced oil recovery in shale. *Energies* 13, 6619 (2020).

36. Lubbers, N. *et al.* Modeling and scale-bridging using machine learning: Nanoconfinement effects in porous media. *Sci. Reports* 10, 1–13 (2020).

37. Santos, J. E. *et al.* Modeling nanoconfinement effects using active learning. *The J. Phys. Chem. C* 124, 22200–22211 (2020).

38. Wang, K. *et al.* A physics-informed and hierarchically regularized data-driven model for predicting fluid flow through porous media. *J. Comput. Phys.* 110526 (2021).

39. Haack, J. R., Hauck, C. D. & Murillo, M. S. Interfacial mixing in high-energy-density matter with a multiphysics kinetic model. *Phys. Rev. E* 96, 063310 (2017).

40. Kadeethum, T. *et al.* A framework for data-driven solution and parameter estimation of pdes using conditional generative adversarial networks. *Nat. Comput. Sci.* 1, 819–829 (2021).

41. Joseph, J. A. *et al.* Physics-driven coarse-grained model for biomolecular phase separation with near-quantitative accuracy. *Nat. Comput. Sci.* 1, 732–743 (2021).

42. Wikipedia. Gaussian process emulator. https://en.wikipedia.org/wiki/Gaussian_process_emulator (2020). Accessed: 2021-07-12.

43. Williams, C. K. & Rasmussen, C. E. *Gaussian processes for machine learning*, vol. 2 (MIT press Cambridge, MA, 2006).

44. Barton, N. R. *et al.* A call to arms for task parallelism in multi-scale materials modeling. *Int. J. for Numer. Methods Eng.* 86, 744–764, DOI: https://doi.org/10.1002/nme.3071 (2011). https://onlinelibrary.wiley.com/doi/pdf/10.1002/nme.3071.

45. Paszke, A. *et al.* Pytorch: An imperative style, high-performance deep learning library. *Adv. neural information processing systems* 32, 8026–8037 (2019).

46. Roehm, D. *et al.* Distributed Database Kriging for Adaptive Sampling ($D^2$KAS). *Comput. Phys. Commun.* 192, 138–147, DOI: 10.1016/j.cpc.2015.03.006 (2015).

47. Rouet-Leduc, B. *et al.* Spatial adaptive sampling in multiscale simulation. *Comput. Phys. Commun.* 185, 1857–1864, DOI: https://doi.org/10.1016/j.cpc.2014.03.011 (2014).

48. Pavel, R. S., McPherson, A. L., Germann, T. C. & Junghans, C. Database assisted distribution to improve fault tolerance for multiphysics applications. In *Proceedings of the 2nd International Workshop on Hardware-Software Co-Design for High Performance Computing*, Co-HPC '15, DOI: 10.1145/2834899.2834908 (Association for Computing Machinery, New York, NY, USA, 2015).

49. Noack, M. *et al.* A kriging-based approach to autonomous experimentation with applications to x-ray scattering. *Sci. Reports* 9, DOI: 10.1038/s41598-019-48114-3 (2019).

50. McKerns, M., Strand, L., Sullivan, T. J., Fang, A. & Aivazis, M. Building a framework for predictive science. In *Proceedings of the 10th Python in Science Conference*, 67–78 (2011). http://arxiv.org/pdf/1202.1056.